\documentclass[twoside,twocolumn]{article}

\usepackage{times}

\usepackage{caption}
\usepackage[pdftex]{graphicx}
\usepackage{algorithmic}
\usepackage{algorithm}

\begin{document}
\pagenumbering{gobble}
%
\title{\textbf{\Large Using GPI-2 for Distributed Memory Paralleliziation of the Caffe 
Toolbox to Speed up Deep Neural Network Training }\\[0.2ex]}

\author{\large Martin Kuehn, Janis Keuper and
Franz-Josef Pfreundt\\[0.3ex]
Competence Center High Performance Computing\\
Fraunhofer Institute for Industrial Mathematics\\
Fraunhofer-Platz 1, D-67663 Kaiserslautern, Germany}

\maketitle

\begin{abstract}
Deep Neural Network (DNN) are currently of great interest in research and application.
The training of these networks is a compute intensive and time consuming task.
To reduce training times to a bearable amount at reasonable cost
we extend the popular Caffe toolbox for DNN with an efficient distributed
memory communication pattern. To achieve good scalability we emphasize
the overlap of computation and communication and prefer fine granular
synchronization patterns over global barriers. To implement these
communication patterns we rely on the the "Global address space Programming Interface" version 2
(GPI-2) communication library.
This interface provides a light-weight set of asynchronous one-sided communication primitives
supplemented by non-blocking fine granular data synchronization mechanisms.
Therefore, CaffeGPI is the name of our parallel version of Caffe.
First benchmarks demonstrate better scaling behavior compared
with other extensions, e.g., the Intel\texttrademark Caffe. Even within a
single symmetric multiprocessing machine with four graphics processing units,
the CaffeGPI scales better than
the standard Caffe toolbox.
These first results demonstrate that the use of standard
High Performance Computing (HPC) hardware
is a valid cost saving approach
to train large DDNs. I/O is an other bottleneck to work with DDN´s in
a standard parallel HPC setting, which we will consider in more detail in a
forthcoming paper.

\end{abstract}

\maketitle

\section{Introduction}

Deep Neural Network (DNN) architectures have improved considerably
the accuracy in data classification opening the door for a plethora
of use cases  in image classification, speech recognition or semantic
text understanding. However, the training of DNNs is a very compute
intensive task. So, the raising interest in these architectures
created a tremendous demand for compute resources which is further 
intensified by a race to greater sizes of DNNs.

Another important factor is the time necessary for training DNNs.
To train a  popular architecture like, e.g., GoogLeNet can easily 
take several days on a Graphics Processing Unit (GPU). 
To make things worse the training usually is
an iterative process of trials and  modifications in the
DNN architecture. So, keeping training times tolerably is a key
requirement to actually apply DNNs in research and industry.

In response to this challenge, hardware vendors brought to market
special hardware, e.g., the DGX-1 sold by NVIDIA or the 
S822LC ("Minsky") sold by IBM. They try to integrate as much compute power
in terms of floating point operations per second (FLOP/s) as possible
in a single compute node. While this special hardware comes also with a
special price it is also not as flexible to apply to other problems
in computer science. On the other hand, there already exists a
plethora of compute systems in the world used for High Performance Computing
 (HPC) \cite{top500}. Usually these consist of hundreds of nodes usually
connected with high bandwidth, low latency networks like InfiniBand
networks. A considerable number of them are even equipped with
GPU accelerators.

\begin{table}[!t]
\caption{APPROXIMATE COMPUTATION TIMES FOR ALEXNET WITH BATCH SIZE $B=256$ AND
450k ITERATIONS, GOOGLENET AND INCEPTION V3 WITH $B=32$
AND 1400k,2000k ITERATIONS.
SETUP CAFFE WITH CUDA 8 AND CUDNN 6
FOR GPUS AND INTEL\texttrademark CAFFE WITH MKL17 FOR CPUS.}
\centering
\begin{tabular}{l|cccc}
 & CPU & K80 & P100 & KNL \\
\hline
AlexNet \cite{krizhevsky2012imagenet}:& & &  & \\
time per iteration & 2s & 0.9s & 0.1s & 0.6s \\
time till convergence & 250h & 112h & 13h & 75h \\
GoogLeNet \cite{szegedy2015going}:& & &  & \\
time per iteration & 1.3s& 0.36s & 0.08s & 0.32s \\
time till convergence & 361h & 100h & 31h  & 89h  \\
Inception V3:& & &  & \\
time per iteration & - &  - & 0.33s & - \\
time till convergence & - & - & 180h  & -  \\
\end{tabular}
\end{table}

Our aim is to make these HPC resources available to the field of
data analytics. The advantage of this approach is twofold. First,
it provides the data analytics community access to the needed hardware quickly
because it is already up and running. Secondly, in the long
run it avoids the separation of resources that are used in the field
of data analytics and in the traditional HPC field. This not only 
simplifies  the buildup and the operation of these compute resources but
it also increases the flexibility to mix data analytics and other 
compute jobs on the same cluster. The latter increases the load factor
and reduces costs.

The toolbox Caffe \cite{jia2014caffe} is very popular to build and train DNNs. It is easy
to use and a wealth of predefined DNNs are available to get to results
quickly. The popular Convolutional Neural Networks usually have
a performance advantage on Caffe versus the TensorFlow framework.
However, the parallelization of Caffe, as it is provided in its original
version, is limited to a single
Symmetric Multi Processing (SMP) compute node.
Intel\texttrademark developed a MPI based prallel version of Caffe, which
we will use as benchmark. 
Our goal is to provide a parallel version of Caffe based on a
Partitioned Global Address Space (PGAS)
Application Programming Interface (API),
that can exploit one sided commmunication more efficiently than MPI.
This speeds up the DNN training
by distributing the computational load on several servers equipped with
a GPU or a set of powerful Central Processing Units (CPUs). 

To ensure high scalability it is
crucial to organize the inter node communication efficiently.
To exploit the precious network bandwidth to the maximum it is
important to overlay as much of the inter node communication
with computation as possible. The
Global address space Programming Interface version 2
(GPI-2) library \cite{gpi2} developed
by our group provides an efficient interface for one sided,
asynchronous data transfers. This interface is the basis of an
efficient and well scaling distributed memory parallelization of 
the Caffe toolbox. So, we call our parallel version of Caffe
CaffeGPI.

The rest of the text is organized as follows. In Section
\ref{s:parallelization} details on our parallelization approach
and communication pattern are given.
In Section \ref{s:implementation} the implementation of the
communication pattern is described. In Section \ref{s:benchmarks}
our first benchmarks are presented and in Section \ref{s:conclusion}
the results are discussed.

\section{Parallelization Approach}

\label{s:parallelization}


Although the numerical operations involved in the training of DNNs are
typically basic linear algebra operations on dense matrices, which have
a rather good FLOP to byte ratio, the bandwidth of the communication networks
must be used efficiently to keep the latency low between the training
iterations.

\subsection{Stochastic Gradient Descend}
The Stochastic Gradient Descend (SGD) \cite{bottou2010large} algorithm is a standard for training
DNN and is implemented in the Caffe toolbox. It is the standard choice for such
famous DNNs like, e.g., GoogLeNet or AlexNet.
The training data $x_i$
is partitioned into batches which are iteratively applied
to the DNN to modify the weights $w$ which are also called the model.
Each iteration consists of a forward
propagation and a backward propagation. In the forward propagation, the data batch is
 inferred while during the backward propagation a gradient on the weights is computed
and later on applied on the model.

\begin{figure}[htbp]
\centering%
\begin{algorithmic}[1]
\REQUIRE $\epsilon>0$
\FORALL{$t=0\dots T$ }
\STATE{\begin{bf}randomly draw\end{bf} batch $M \leftarrow B$ samples from $X$}
\STATE{\begin{bf}Init\end{bf}${\bf\Delta}w_t=0$ }
\FORALL{$x\in M$ }
\STATE{\begin{bf}aggregate update\end{bf} ${\bf\Delta}w \leftarrow \partial_wx_j(w_{t})$}
\ENDFOR
\STATE{\begin{bf}update\end{bf} $w_{t+1} \leftarrow w_{t} - \epsilon{\bf\Delta}w_t$}
\ENDFOR
\STATE Return $w_T$
\end{algorithmic}
\caption{Mini-Batch SGD
with samples $X=\{x_0,\dots,x_m\}$, iterations $T$, step-size
$\epsilon$, batch size $B$ \label{fig:algo_mini}}
\end{figure}

The two basic parallelization strategies commonly used for SGD
are data parallelism or model parallelism \cite{KeuperDL}.
In the model parallelism, the net and its weights are distributed among different ranks
of the parallel computer, while the batch stays the same for all ranks.
In data parallelism the DNN is duplicated on every rank while the data batch
is split into equal fractions for each rank. The computed gradients on each rank
are aggregated and then applied to the model which is distributed back to all the ranks.
We concentrate on the data parallelism approach here which is favorable for
DNN containing many convolutional layers like the popular AlexNet or GoogLeNet.
An advantage of this approach is a constant aggregated IO turnover over
the ranks fetching the training data.

As the DNN is organized in layers we write the model of layer $l$ in
 the iteration $k$ as $w^{(l,k)}$, the partial gradient on rank $r$ as
$\Delta w^{(l,k)}_r$. Using this notation the iterative data parallel 
SGD algorithm can be shortly noted as

\begin{equation}
w^{l,k+1} :=  w^{l,k} -  \epsilon \sum_{r=1}^s\Delta w^{l,k}_r.
\end{equation}

Higher order terms are neglected here without loss of generality. These terms
usually depend on the history of the models $w^{l,k}$,
which are inherently broadcasted
to all the ranks to perform the next forward propagation phase.


\subsection{Design Principles}

\begin{figure}[htbp]
\centering%
\includegraphics[width=\linewidth ]{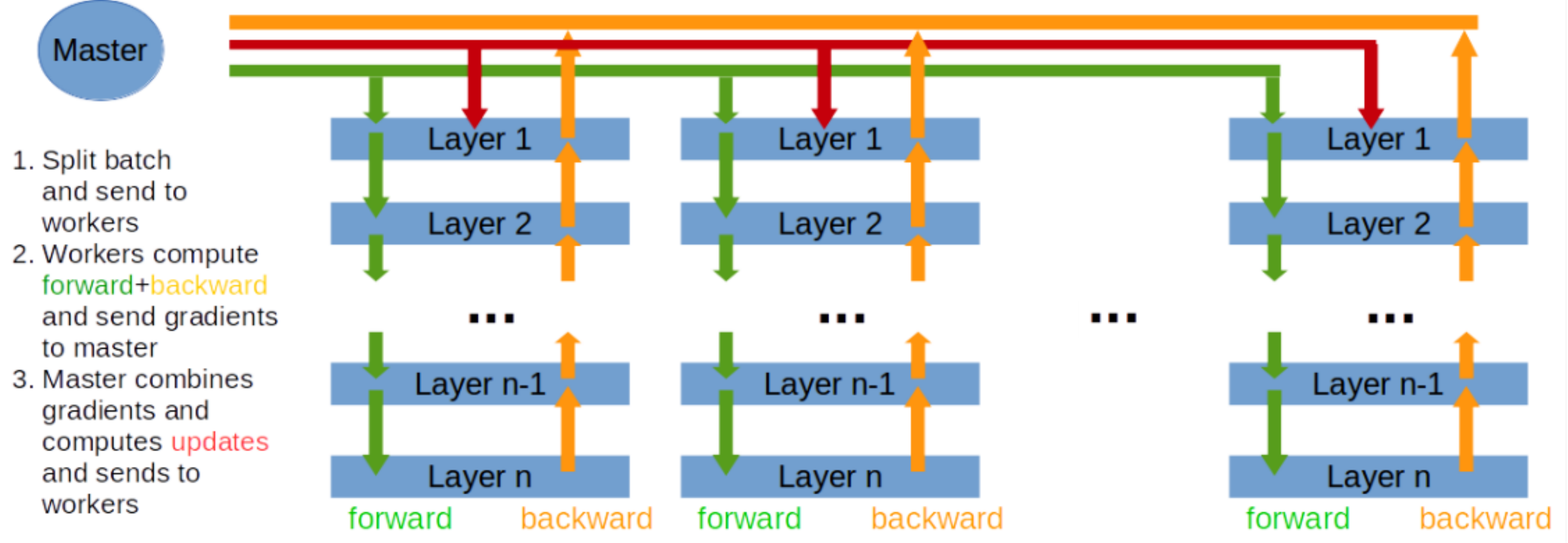}
\caption{Data Parallel SGD as frequently implemented.}\label{fig:traditionalParallelSGD}
\end{figure}

To exploit the full potential of this parallelization
approach the data transfers of the model
and gradients have to be designed very carefully, since
the total amount of data that has to be transferred during each iteration
grows with the number of compute ranks.

As a consequence a typical HPC interconnect, like EDR InfiniBand, should
be used to provide enough bandwidth to exploit the scalability of the problem.
These networks are commonly used to connect the nodes in
current HPC clusters.
Another important aspect is to use these interconnects efficiently. Foremost
this means to overlap computation and communication to avoid that
their run times add up fully. Instead both
should ideally take place at the same time so that no additional time for communication
is necessary. Unfortunately this is not the case for the standard Caffe which
enters separate phases for computation and communication (see Figure \ref{fig:traditionalParallelSGD}).
The second principle is to avoid global synchronization points
between the ranks. Instead data dependencies are enforced locally
and very fine granular. The communicated data chunks are pipelined to keep the ranks
busy most of the time with either computing, communication or both.
Barriers to synchronize the ranks are avoided and used only where
absolutely necessary.
However, it is important to note that our approach strictly regards all the data
dependencies immanent of the SGD algorithm unlike the so called "asynchronous SGD"
algorithms described in literature \cite{ASGD}.

\subsection{Overlapping Computation and Communication}
\label{ch:communication}

\begin{figure}[htbp]
\centering%
\includegraphics[width=\linewidth]{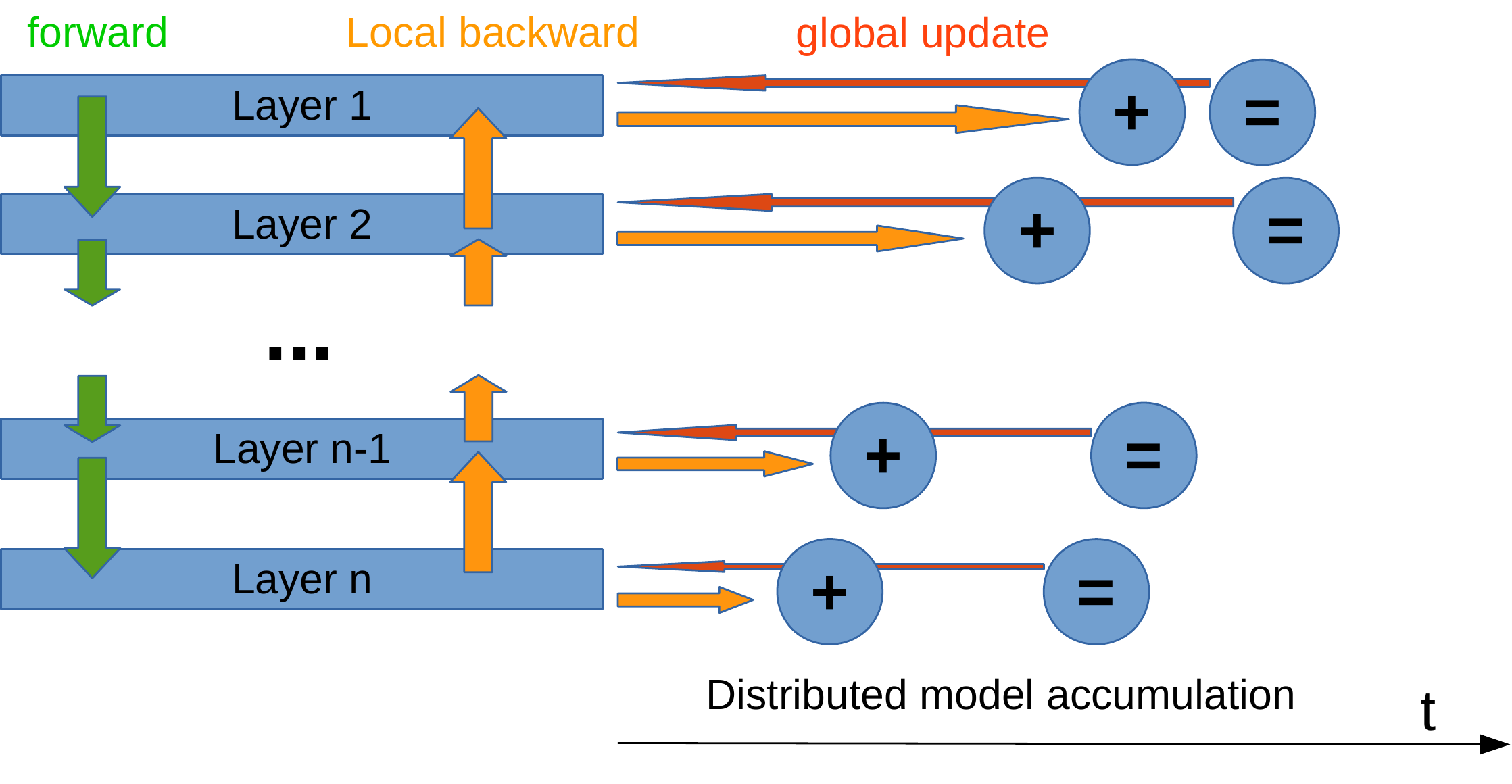}
\caption{Sketch of the data parallel SGD communication pattern implemented in this work.}
\label{fig:ourSGD}
\end{figure}

During the backward propagation the partial gradients $\Delta w^{(l,k)}_r$
are computed separately for each layer and in an inverse order.
The next read access to the model of a certain layer is in the forward propagation of
the following time step. So, especially for the layers at the bottom of the DNN
quite some time is available to reduce the partial gradients and to
update the model of that layer (see Figure \ref{fig:ourSGD} for illustration).

The communication pattern is turn based, one layer of the DNN per turn.
In each turn, a partial gradient $\Delta w^{(l,k)}_r$ is computed for the
respective layer $l_i$
and  forwarded to the receiving ranks.  Incoming partial gradients of previous layers from other
ranks are checked, aggregated if available and forwarded to receiving ranks 
as well.
In the same manner, updates on the model of previous layers are forwarded to
the receiving nodes. In all these cases the data transfers are only triggered but not
awaited for conclusion.

\begin{figure}[htbp]
\centering%
\begin{algorithmic}[1]
\FORALL{$l=L,\dots, 1$ }
\STATE{\begin{bf}compute\end{bf} local gradient $\Delta w^{(k,l)}_r$ }
\STATE{\begin{bf}check\end{bf} for arrived gradients from previous layers}
\STATE{\begin{bf}reduce\end{bf} arrived gradient data locally}
\STATE{\begin{bf}trigger\end{bf} sends of available gradient data}
\STATE{\begin{bf}trigger\end{bf} sends of arrived model data}

\ENDFOR
\STATE{\begin{bf}finalize\end{bf} local communication phase}

\end{algorithmic}
\caption{Turn based communication pattern on rank r.}
\label{fig:algo_communication}
\end{figure}

In the finalization phase, all loose ends of the communication
to the local rank $r$ are
finalized. After that the local instance of the DNN is ready
for the next iteration. Please note that even between  iterations
there is no global barrier applied. So, every rank that received
a complete update of the model can immediately start with the next training
phase without having to wait for other ranks to get their full
update.

Particularly, this scheme allows to overlap the computation of the
gradient $\Delta w^{(l,k)}_r$ with the communication of gradients and models of
previous layers. And it avoids global barriers between the ranks.

\section{Implementation}

\label{s:implementation}

Our parallel version of the Caffe framework is done as minimally invasive
as possible. Basically, the setup routine of the DDN and the backward
propagation routines are modified. Additionally, a layer-wise model update
is introduced in the Solver class.

During the backward propagation over the layers, the calculations
of the gradients are followed by calls to the newly added communication 
routines to reduce the local gradients and to broadcast the updated model.

\subsection{Basics of GPI-2}


To implement the overlapping, one sided communication pattern the GPI-2 library
has been used, which is a PGAS
communication API for C/C++ and Fortran applications. Fulfilling the
Global Address Space Programming Interface (GASPI) specification (see webpage \cite{gaspi}),
it provides truly asynchronous one-sided communication primitives supplemented by
a non-blocking light-weight and fine granular data synchronization mechanism.
GPI-2 exploits interconnects supporting
Remote Direct Memory Access (RDMA) as, e.g., InfininBand networks.
On these networks the data transfers can be almost completely offloaded
to the network infrastructure reducing the load on the computational resources
to a minimum. No intermediate copies are necessary which saves memory
bandwidth. Apart from that, GPI-2 is a very lean library and gives
the user more direct control over the particular data transfers as,
e.g. the usual Message Passing Interface (MPI) library.

All these features make GPI-2 a perfect match to implement the
overlap of computation and inter node communication in Caffe. Being an open source
library GPI-2 can be downloaded at \cite{gpi2}.

\subsection{Implementation of Data Transfers}

In the Caffe data structures that define the DNN the arrays that carry
the model and the gradient data are placed inside  GPI-2 data segments.
Providing the gaspi\_segment\_use function, GPI-2 cooperates
perfectly with special memory regions in Caffe which are 
allocated by cudaMallocHost to enhance data transfers to the GPU.
The GPI-2 library allows to write remotely (inter node) and directly  to these
segments.
All the data transfers are triggered with a gaspi\_write\_notify
call to the library. The receiver of the data chunk checks on the
respective notification and acts on the received data if necessary.

The gradient data is reduced in a reduction tree pattern aggregating
the final gradient on the master rank. The master rank performs the
update on the current model to compute an update for the next iteration.
Then the updated model is broadcasted to all the other ranks in another
tree pattern. The gradient data is always sent from higher rank numbers to lower rank numbers
while the updated model is broadcasted from lower rank
numbers to higher rank numbers. As the size of the gradients equals the size of
the models we take advantage of the full duplex feature of switched networks.

The reduction and the broadcast trees are build from scratch to
keep control over the tree topolgy and to interwine
closely the communication pattern with the computation.

\subsubsection{Reduction of the Gradient}
To reduce the local gradients in a binomial, tree each rank checks for incoming
gradient data in its receive buffer.
If available the gradient data is reduced (added to) with its own
gradient data of that layer. If the rank has a receiver in the tree pattern
the reduced gradient is forwarded to this receiver using a call
to  gaspi\_write\_notify. The communication
tasks are performed once in the loop over the layers as 
depicted in section \ref{ch:communication}. The gradient data is processed
as available. No waiting takes place for a specific data chunk.

\subsubsection{Broadcast of the model}
The broadcast of the model is performed similarly as the gradient reduction.
As no reduction steps are necessary the incoming model data from the sender is
just forwarded to its receiver ranks using a call to gaspi\_write\_notify.


\section{First Benchmarks}
\label{s:benchmarks}

\begin{figure}[htbp]
\centering%
\includegraphics[width=\linewidth]{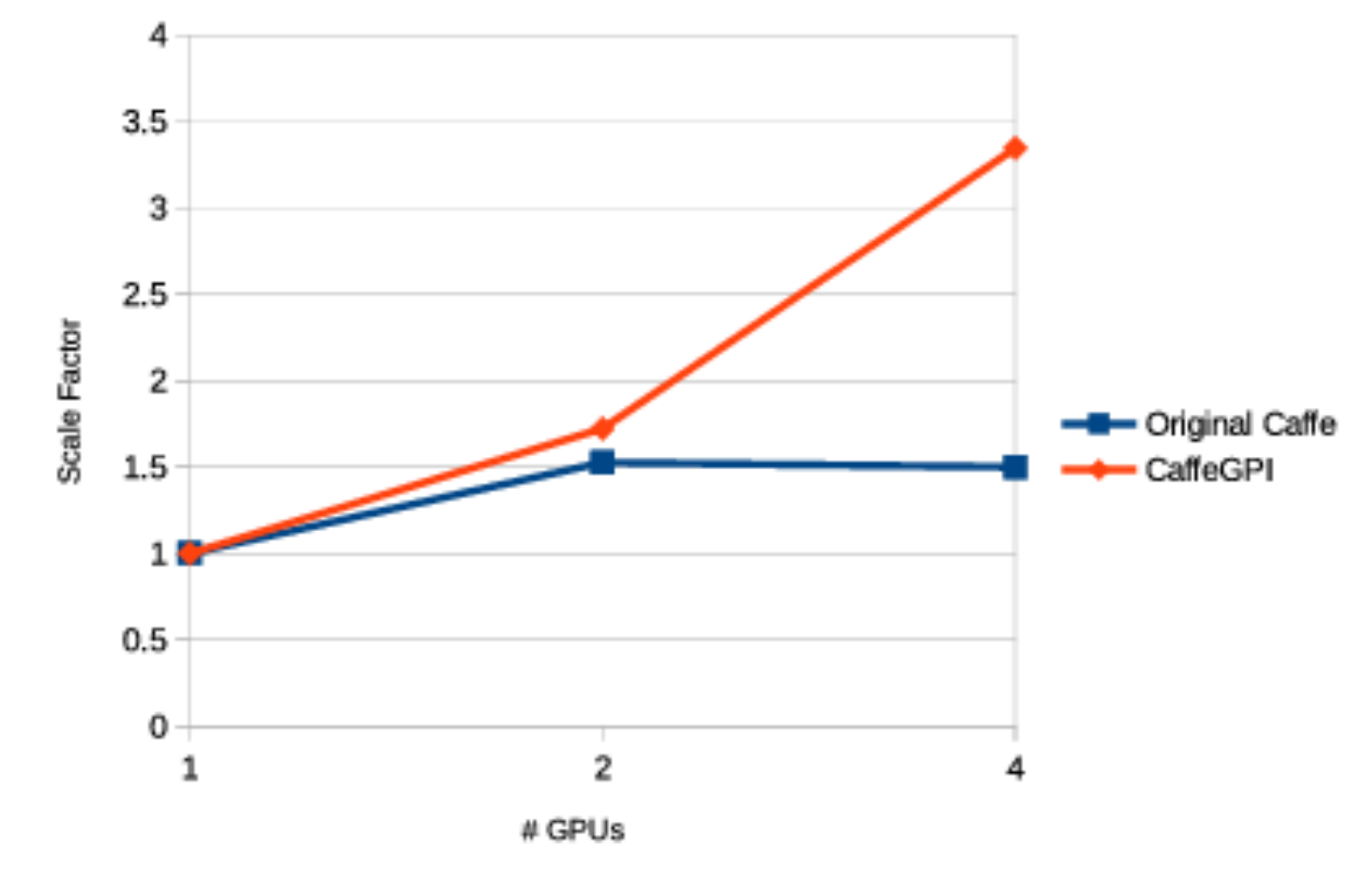}
\caption{Scaling results for AlexNet an overall  batch size of $256$ on a single node with multiple GPUs
interconnected via PCIe. Based on Caffe 1.15, Cuda 8, CuDNN 5.1.  }
\label{fig:ResAlexSingle}
\end{figure}

To evaluate our parallelization approach we start to compare CaffeGPI with
the original Caffe on a SMP machine containing 4 GPUs. This setup is quite
similar to specialized workstations produced to train DNNs.
The original Caffe
uses the standard thread parallel communication pattern in this benchmark.
In the CaffeGPI benchmark,
4 independent processes are started on the same node, one for each GPU,
communicating through the network card.
The SMP machine is a single
node of the Taurus cluster at the ZIH in Dresden containing 2 Intel\texttrademark
 Xeon E5-2680v3 CPUs, 64GB
Random-access memory (RAM) and
4 NVIDIA K80 GPU. 
The network is InfiniBand FDR. As DNN we choose the familiar AlexNet.
Figure \ref{fig:ResAlexSingle} depicts the scaling behavior at 1, 2, and 4 GPU.

\begin{figure}[htbp]
\centering%
\includegraphics[width=\linewidth]{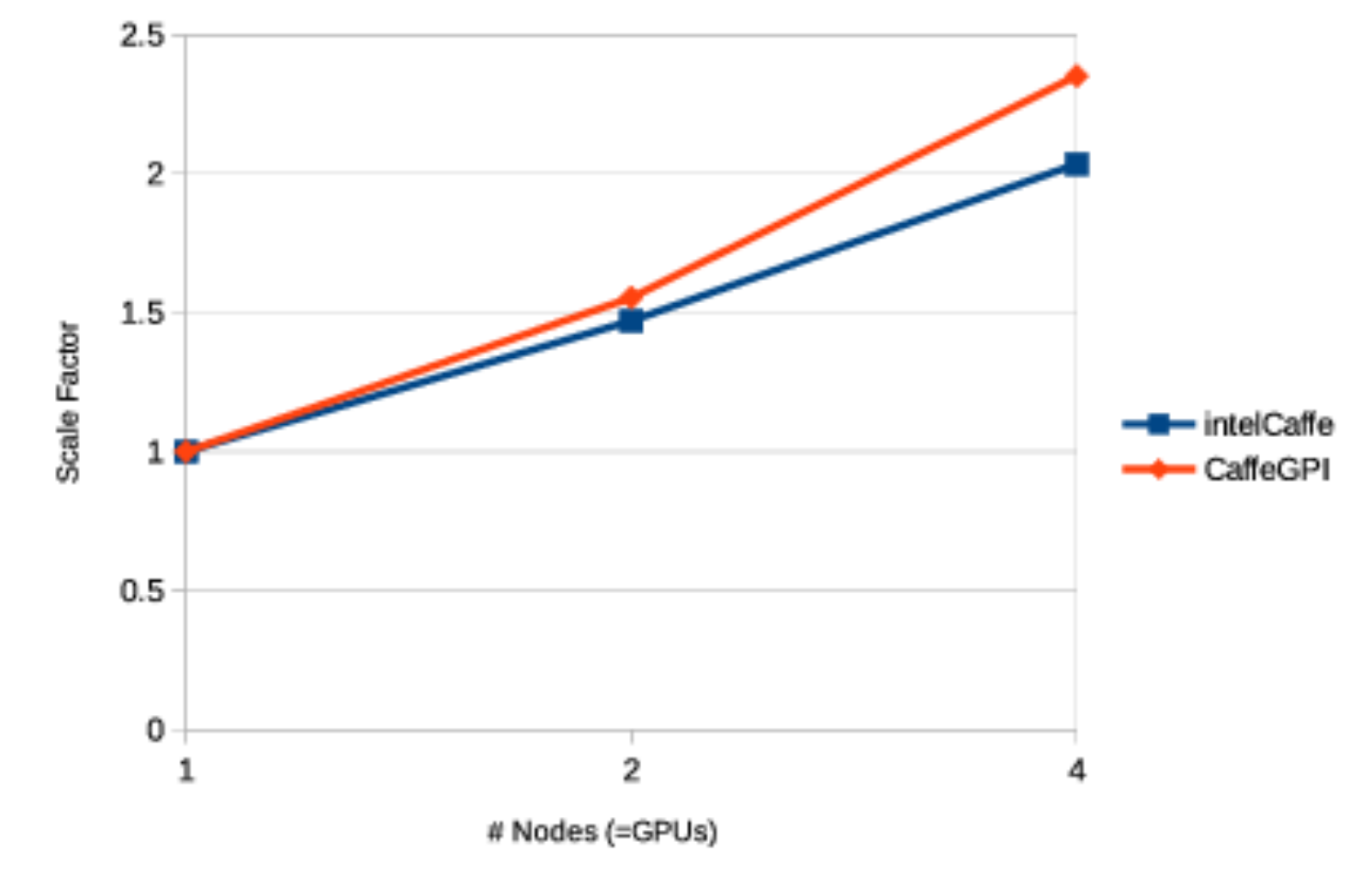}
\caption{Scaling results for AlexNet with an overall  batch size of $256$ on distributed nodes with single GPUs interconnected via Infiniband. Based on Intel\texttrademark Caffe 1.14, Cuda 8, CuDNN 5.1}
\label{fig:ResAlexDist}
\end{figure}

On a first glance our CaffeGPI should have a small disadvantage in this benchmark
compared with the standard Caffe toolbox because it
needs to communicate via memory copies inside the node. At a closer look
the reduction of the partial gradients puts a lot of load on the memory system,
the PCIe bus and the QuickPath Interconnect between the CPUs. In both cases, 
memory copies between GPU-RAM and CPU-RAM have to be performed.
However in the benchmark of the  standard Caffe all the GPUs execute their communication phase
at the same time leaving precious bandwidth idle during the computation phase.
In the CaffeGPI benchmark, the memory copies between GPU-RAM and CPU-RAM are not
overlapped but interleaved with computation of the partial gradients. As not all the 
GPUs perform their copies at the same time, the memory transfers are distributed
over a longer period of time. The memory copies 
across the two sockets are performed by the network card and overlapped with the computation.
Finally our implementation can
demonstrate superior scaling behavior as depicted in Figure \ref{fig:ResAlexSingle}.
The second benchmark compares CaffeGPI to Intel\texttrademark Caffe (see webpage \cite{intelcaffe}),
a distributed memory
extension of Caffe based on the MPI.
In this benchmark, 1, 2 or 4 nodes
of the same cluster were used, but only one GPU per node. Here distributed memory
data transfers are performed in both scenarios.
The benchmark in Figure \ref{fig:ResAlexDist}
 demonstrates a superior scaling behavior of our implementation
in comparison to the Intel\texttrademark Caffe framework. A  speedup of 2.4 on 4 nodes compared
to one node delivers a reasonable  performance figure to train AlexNet in a
reasonable time frame on standard HPC hardware.

\section{Conclusion and Future Work}
\label{s:conclusion}

The preliminary benchmarks presented in this work demonstrate that our
distributed memory communication pattern implemented in the CaffeGPI framework
scales well on four distributed memory nodes equipped with one GPU per
node. The total performance is similar or even better than
using the standard SMP-parallel approach of Caffe on a SMP node equipped 
with 4  GPU.
Even on this single SMP node with 4 GPUs, our CaffeGPI scales much better
than the standard Caffe framework.

These results demonstrate that data scientists can rely on available HPC
compute resources to train their DNNs in a reasonable time frame. Our toolbox CaffeGPI
can help to satisfy the need for more compute power in the area of data science
without having to buy vast amounts of specialized hardware, which is difficult
to apply economically for other tasks in computer science.

We will continue to benchmark various hardware configurations,
e.g.  a NVIDIA DGX-1 or an IBM S822LC ("Minsky").
Further benchmarks will be done to
analyze the communication pattern introduced
in CaffeGPI. Alternative patterns will be evaluated
that might improve the reduction and the broadcast operations.
We will also extend our benchmarks on more
DNNs and to wider batch sizes to evaluate their scaling behavior and to find performance
optimized training parameters.


\section*{Acknowledgment}
\noindent The authors thank the Center for Information Services and High Performance Computing (ZIH) at TU
Dresden for generous allocations of computer time. We also gratefully acknowledge the support of NVIDIA Corporation 
with the donation of the Titan X Pascal GPU used for this research.

\bibliographystyle{IEEEtran}
\bibliography{iariacite}

\end{document}